# A Survey on Graph Neural Networks for Fraud Detection in Ride Hailing Platforms

Kanishka Hewageegana*
*School of Computing*
*Informatics Institute of Technology*
Colombo, Sri Lanka
kanishka.20201000@iit.ac.lk

Janani Harischandra
*School of Computing*
*Informatics Institute of Technology*
Colombo, Sri Lanka
janani.h@iit.ac.lk

Nipuna Senanayake
*School of Computing*
*Informatics Institute of Technology*
Colombo, Sri Lanka
nipuna.s@iit.ac.lk

Gihan Danansuriya
*Faculty of Applied Sciences*
*Rajarata University,*
Sri Lanka
gihanprabudda@gmail.com

Kavindu Hapuarachchi
*Faculty of Applied Sciences*
*University of Sri Jayewardenepura,*
Sri Lanka
kavinduhapuarachchi@gmail.com

Pooja Illangarathne
*School of Computing*
*Informatics Institute of Technology*
Colombo, Sri Lanka
pooja.20210435@iit.ac.lk

*Abstract*—This study investigates fraud detection in ride-hailing platforms through Graph Neural Networks (GNNs), focusing on the effectiveness of various models. By analyzing prevalent fraudulent activities, the research highlights and compares the existing work related to fraud detection which can be useful when addressing fraudulent incidents within the online ride hailing platforms. Also, the paper highlights addressing class imbalance and fraudulent camouflage. It also outlines a structured overview of GNN architectures and methodologies applied to anomaly detection, identifying significant methodological progress and gaps. The paper calls for further exploration into real-world applicability and technical improvements to enhance fraud detection strategies in the rapidly evolving ride-hailing industry.

## I. INTRODUCTION

The landscape of fraud detection has undergone significant transformations in recent years, specially in the aftermath of the COVID-19 pandemic. The rise in digital transactions and the proliferation of ride-hailing platforms have led to an increase in fraudulent activities, necessitating advanced detection mechanisms [28]. Traditional methods of fraud detection have relied on a variety of network types, including simple statistical models and machine learning algorithms, which analyze transactional data for anomalies. Unfortunately, these methods often fall short in capturing the complex and dynamic interactions between users and transactions specially in modern ride-hailing ecosystems. This sheds light on the exploration of deep learning approaches to incorporate fraudulent incidents.

The appearance of GNNs has marked a pivotal shift in fraud detection methodologies by leveraging the relational information present in data, making them particularly adept at identifying fraudulent activities within ride-hailing platforms [12],[13]. These networks excel in aggregating neighborhood information and uncovering subtle patterns of fraudulent behavior, which are often camouflaged within legitimate transactions.This is particularly relevant in the context of social networks, where entities (passengers, drivers and trips) and their interactions form a complex web of relationships. GNNs can capture the relational dependencies and patterns within this data, enabling a more sophisticated and effective detection of fraudulent activities [22].

This paper presents a thorough analysis of the popular fraudulent types in the ride-hailing ecosystem, foundation to GNNs, intersection of Fraud Detection with GNNs, Dynamic and Static Concepts and their impact is presented. The rest of this paper is organized as follows; Related Work, Analysis of Existing Work, Summarized Taxonomy, Results and Discussion and finally provides a Conclusion of the review paper.

## II. TYPES OF TRENDING FRAUDS WITHIN THE RIDE HAILING ECOSYSTEM

This section explores different kinds of dishonest activities within the ride-hailing sector, such as tampering with GPS data, taking unnecessary long routes, conspiring between drivers and passengers to fake rides, and negotiating fares off the books. These schemes pose a threat to the ride-hailing companies' revenue and could endanger passenger safety and the overall trustworthiness of the service. Based on the comprehensive summary, the research community will be able to understand the current status quo of the industry and will be able to start their efforts based on that.

## A. Fake GPS Systems

One type of fraud involves drivers manipulating the system to chase targets or bonuses, particularly during challenging times like a pandemic. In Indonesia, drivers have been found to cheat to meet targets set by ride-hailing companies, driven by competition and the desire to earn their relavent bonus incentives [15]. This behavior is partly attributed to the gamification strategies employed by these companies, which create an illusion of freedom and incentivize drivers to maximize their earnings, sometimes through fraudulent means.

To achieve these targets, some driver's resort to using fake GPS applications to manipulate their location. According to [15] by gaining root access or jailbreaking their smartphones, they can trick the system into believing they are closer to potential passengers, thereby increasing their chances of receiving ride requests. This method not only gives them an unfair advantage over other drivers but also leads to longer wait times for passengers, as the driver is actually farther away than the system indicates which leads to major concerns on the trustworthiness of the service.

## B. Route Manipulation and Long-Hauling in Ride-Hailing Services

Within the Ride-Hailing Platforms,taxi trajectory anomalies represent a sophisticated form of fraud. This type of fraud involves drivers deviating from standard or optimal routes for various deceptive purposes. The technical approach to this fraud typically involves drivers deliberately taking longer or less efficient routes than necessary. This can be done to artificially increase the fare by extending the travel distance or time, a practice known as **"long-hauling".** In some cases, drivers might also engage in more deceptive practices, such as coordinating with others to create traffic patterns that justify longer routes or manipulating the ride-hailing app's GPS data to create a false representation of the route taken [20].

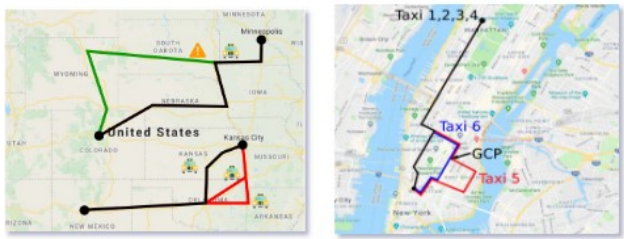

Fig. 1. Group of taxi trajectory fraud [20].

## C. GPS Spoofing

Drivers use GPS Spoofing Applications to manipulate their geographic location. By altering their GPS data, they can make it appear as though they are picking up and dropping off passengers at different locations than they actually are. This allows them to create fictitious rides, logging trips that never actually took place, and thereby fraudulently increasing their earnings [23]. These dishonest activities include creating fake rides in order to receive incentives or organizing rides and drivers to create fake ride histories in order to unfairly profit from the system's incentive proceedings. This phenomena is not limited to any single region; rather, it is present in many nations, especially those that offer attractive incentives for ride-hailing services.

## D. Ride Collusion

This technique involves collusion between a driver and a fake passenger (which could be the driver using a second account or an accomplice). The driver and the colluding passenger coordinate to book rides that are never actually taken. The passenger requests a ride and the driver accepts it; however, no actual transportation occurs. The ride is completed on the app as if it were a legitimate trip, resulting in the driver receiving payment for a service that was never provided [23]. This practice artificially inflates their ride numbers, helping them meet company targets or earn bonuses. However, if another driver receives the request, they simply cancel it.

## E. Hire Conversions and Trip Manipulations

Fraudulent activities in the ride-hailing sector, particularly prevalent in countries like Sri Lanka, pose a significant threat not only tos the financial stability of companies such as PickMe and Uber but also to passenger safety and trust. A common tactic observed involves drivers manipulating the hiring process, known as **'Hire Conversions'** and **'Premature Trip Conversions (PTC)'**

In this particular scenario, drivers, upon reaching the customers' pickup location, engage in direct negotiation with the customer, offering to complete the trip at a reduced fare compared to the one quoted in the app at the pickup point itself or offering after traveling for some distance. This practice, primarily motivated by the desire to circumvent the commission fees charged by the ride-hailing service, presents an attractive proposition to customers due to the lower cost. Consequently, customers often agree to these terms. Following this agreement, the driver may choose to go offline by clicking the complete button on the ride-hailing application, thereby obscuring the trip from the platform's oversight, and proceed to transport the customer to their destination.

Such practices not only undermine the revenue model of the ride-hailing platforms but also raise concerns about the traceability and accountability of the trips, which are essential for ensuring passenger safety and maintaining service reliability. This trend highlights the need for stricter enforcement of platform policies and the development of more robust systems to detect and prevent these unauthorized transactions.

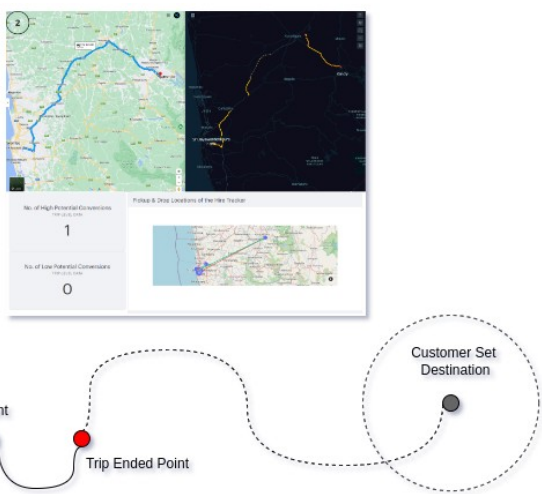

Fig. 2. Hire Conversion Stages

III. FOUNDATION TO GRAPH NEURAL NETWORKS (GNN)

GNNs have emerged as a powerful framework for learning representations of graph-structured data, enabling the application of machine learning to complex systems such as social networks, molecular structures, and, notably, fraud detection mechanisms. At the core of GNNs is the concept of message passing, which allows for the aggregation and transformation of information from a node's neighbors to generate node-level, edge-level, or graph-level representations.

*A. Basic Graph Structure*

Formally, a graph G is defined as an ordered pair G = (V, E), where V is a set of vertices or nodes, and E is a set of edges, each linking a pair of nodes. Nodes represent discrete entities in a graph, which can be anything from data points in a network analysis to cities in a route map. They are the fundamental units from which graphs are constructed. Nodes are endowed with a set of features X= {$x_1$, $x_2$, $x_3$, ...}, where each feature, such as $x_1$, denotes a unique attribute. These attributes can encode various properties such as numerical values, categorical labels, or even high-dimensional vectors obtained from complex data representations.

Edges (or links) represent the connections or relationships between pairs of nodes. The nature of these relationships can vary widely; for instance, in social networks, they might represent friendships, while in molecular structures, they might denote chemical bonds. The existence of an edge between nodes i and j is denoted by $(i,j) \in$ E.

Node features are inherent attributes associated with the nodes, providing a multi-dimensional characterization of each node. These features are crucial for tasks that involve node classification, clustering, or regression within graph-based machine learning models. For example, in a hire conversion trip manipulation graph, the node features might include trip's id, number of trips, and other trip level information.

GNNs are a specialized subset of Artificial Neural Networks (ANNs) tailored for processing data represented in graph form. This capability renders them highly effective for tasks such as node classification, link prediction, and graph classification. The representation of the edges and nodes of a graph in a digestible format by a neuron in an ANN involves the construction of an adjacency matrix, **A**. The adjacency matrix, is a binary square matrix of size |V|×|V| used to represent the connectivity structure of a graph, offers a compact representation of the graph structure in computational graph theory. If there is an edge connecting nodes i and j where $1 \leq i, j \leq n$, an entry $A_{ij}$ in the adjacency matrix is 1, and 0 otherwise. This adjacency matrix contains multiple advantages; if the matrix is symmetrical for undirected graphs, which implies that $A_{ij} = A_{ji}$. For weighted graphs, the adjacency matrix contains weights instead of binary values, quantifying the strength or capacity of the connections

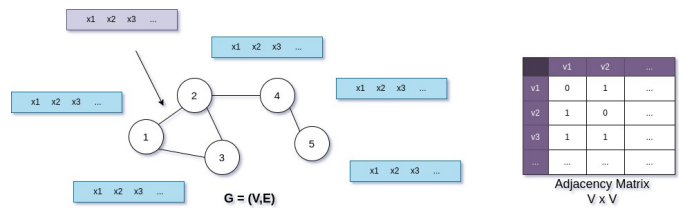

Fig. 3. Basic Graph Structure

*B. Graph Data Characteristics*

Graphs are inherently distinct from traditional data structures, which is evident when comparing their fundamental characteristics. Understanding of these differences is very important when effectively applying graph theory to real-world problems.

*a) Size Independent:* Graphs are inherently size-independent and can represent structures of varying scales, from small molecule compounds to massive networks. This flexibility stands in contrast to array-based data structures that are size-dependent. For example, under image analysis, a pixel grid has a fixed resolution, whereas a graph structure can adapt to represent the same image at any resolution, abstracting the content regardless of scale.

*b) Isomorphism and Permutation Invariance:* Graphs embody the concept of isomorphism, where different graphs may represent the same structure if there is a bijection between the vertex sets of the two graphs that preserves the edge relation. This translates to a key property known as permutation invariance. In a machine learning context, this means that the input features to a graph neural network can be the rows of the adjacency matrix regardless of their order, as the underlying structure remains unchanged. This characteristic is crucial when learning from graph data, as the model should be insensitive to the order of nodes.

*c) Grid Structure and Non-Euclidean Space:* Graph data doesn't fit into the Euclidean geometry, which are straight-line patterns we're used to with normal shapes and grids. This is because it exists in a more complex space that doesn't follow the simple rules of shapes like squares and circles. Instead, graph data is made up of points (nodes) and the lines connecting them (edges), creating its own unique layout. This setup is great for showing complicated relationships and structures, like how people are connected in social networks, the way molecules are structured in chemistry, or how routes link up in transport systems.

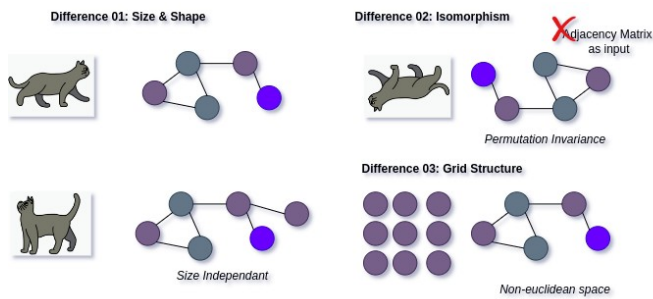

Fig. 4. Graph Data Characteristics and Differences

### C. Categorization of Graph Analytical Problems

Graph analytical problems can be categorized based on the level at which predictions are made: node, edge, sub-graph, and graph levels.

*a) Node-Level Prediction:* This involves predicting properties or labels for individual nodes within a graph. For instance, given a social network graph, one might want to predict whether a particular user (an unlabeled node) smokes. The prediction is based on the user's attributes and their connections within the graph. This type of problem is foundational in personalized recommendation systems and user profiling.

*b) Edge-Level Prediction (Link Prediction):* Edge-level predictions are concerned with the likelihood of the existence of a link between two nodes. A classic application is predicting the next movie a Netflix user might watch, based on their viewing history and the viewing histories of similar users. By analyzing the existing links and patterns within the graph, algorithms can infer potential new connections, which has significant applications in social network analysis and recommendation engines.

*c) Sub-Graph Level Prediction (Community Detection):* This category involves identifying clusters or communities within a graph, which is analogous to detecting social circles within a larger social network. The goal is to determine whether a set of nodes forms a community. This problem is crucial for understanding the modular structure of networks, which can have implications for marketing strategies and information dissemination analysis.

*d) Graph-Level Prediction:* Here, the task is to predict properties or labels for entire graphs or sub-graphs. A typical application is in the field of chem-informatics, where one might predict whether a molecular structure is likely to be an effective drug. This involves analyzing the entire graph representation of the molecule to assess its properties or activity.

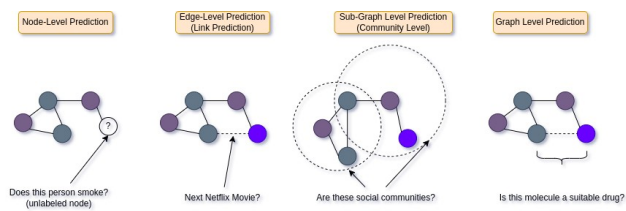

Fig. 5. Types of Graph Problems

### D. Types of Graph Neural Networks

This section will focus on the main types of GNNs that are relevant to the ride-hailing ecosystem, as well as how to address the class imbalance among the various types, which is a major issue in fraud detection.

*a) Graph Convolutional Networks (GCNs):*

These architectures are among the most popular and widely utilized GNNs in the literature, effectively extending the principles of convolutional neural networks to graph structures [53]. These networks adeptly manage the irregular structure of graphs by aggregating and transforming features from a node's neighborhood. A notable characteristic of GCNs is their stability under stochastic perturbations, indicating their resilience to noise or randomness in the data. This stability presents a considerable advantage across various applications. A study utilizing a new RES and a GFT model over GCNs to assess the robustness of the graph filter discovered that GCNs maintain stability in the face of stochastic perturbations, with a factor proportional to the link loss probability [53]. However, it has been observed that increasing the width and depth of a GCN can decrease its stability while enhancing its performance.

Within the Context of fraud detection, class imbalance is a common issue hence there are less number of frauds occur when compered to the other normal incidents. Hence, In addressing the challenges posed by imbalanced datasets, GCN can be adapted to use a weighted cross-entropy loss function. This adaptation is crucial for enhancing the model's sensitivity towards minority classes, which are often of greater interest [24],[9]. The weighted cross-entropy loss function for a binary classification task in the context of GCNs can be expressed as:

$$L = \frac{-1}{N}\sum_{i=1}^{N}[w_1 . y_i . \log(p_i) + w_0 . (1-y_i) . \log(1-p_i)]$$

Where;
- $L$ : Loss
- $N$ : number of samples
- $y_i$ : true label of the *i*-th sample
- $p_i$ : predicted probability of the *i*-th sample belongings to the positive class.
- $w_1$ and $w_0$ are the weights assigned to the positive and negative classes, respectively, designed to counteract the class imbalance by assigning a higher weight to the minority class.

The calculation of this loss involves first performing a forward pass through the GCN to obtain the predicted probabilities $p_i$ for each node (or graph, depending on the task). These probabilities are then used in conjunction with the true labels $y_i$ to compute the loss $L$ using the above formula.

During training, the model's parameters are updated to minimize $L$, typically using gradient descent-based optimization algorithms.

### b) Graph Attention Networks (GATs):

GATs represent a novel approach by enabling nodes to dynamically assign different levels of importance to their neighbors without the need for costly matrix operations or prior knowledge of the graph structure, introduced by [5]. This is very interesting because GATs leverage self-attention mechanisms—a concept borrowed from sequence models—to weigh the influence of nodes during feature aggregation. Mathematically, the attention coefficients, $\alpha_{ij}$, that a node $i$ assigns to its neighbor $j$ are computed as a function of their features, $h_i$ and $h_j$, typically through a softmax over all neighbors $N_i$ of node $i$:

$$\alpha_{ij} = softmax_j(e_{ij}) = \frac{\exp(e_{ij})}{\sum_{(k \in N_i)} \exp(e_{ik})}$$

where $e_{ij}$ is a learnable function of the features of i and j, often implemented as a single-layer feed-forward neural network. This mechanism allows GATs to model complex, node-dependent relationships within the graph, enhancing their applicability to a wide range of tasks, including fraud detection.

The cross-entropy loss function is use in the GAT which is a popular metric for multi-class classification tasks because it can be used to measure how well a classification model performs when the output is a probability value between 0 and 1. Cross-entropy loss is a great option for training classification models, especially those that handle class imbalance, since it increases as the predicted probability diverges from the actual label. The following is the formula for multi-class cross-entropy loss:

$$L = -\sum_{i=1}^{N} \sum_{c=1}^{C} y_{i,c} \log(p_{i,c})$$

Where;
- $L$ : Loss
- $N$ : Number of samples
- $C$ : Number of classes
- $y_{i,c}$ : Binary indicator of whether class *c* is the correct classification for observation *I*
- $p_{i,c}$ : Predicted probability that observation *i,* belongs to class *c*

To calculate the multi-class cross-entropy loss, begins by obtaining the model's predicted probabilities for each class for every sample. These predictions are then compared to the actual class labels, which represented in a one-hot encoded format. For each sample, the loss is computed by taking the negative logarithm of the predicted probability corresponding to the true class and summing this across all classes. However, only the true class contributes to the sum due to the one-hot encoding. The total loss is then derived by averaging these individual losses across all samples in the dataset. To address class imbalance, weights can be introduced to the loss calculation, assigning greater importance to the minority classes. This weighted approach helps to ensure that the model does not overlook the less frequent classes, thereby improving its overall predictive accuracy and fairness.

To tackle class imbalance with the cross-entropy loss, modifications such as logit adjustment techniques can be applied. [44] introduces modifications to standard cross-entropy loss functions by extending a logit adjustment technique to cope with class imbalance, demonstrating the approach's validity through superior performance compared to competing methods in the context of linear B-cell epitope prediction and classification.

### c) Graph Isomorphism Networks (GINs):

This architecture mainly helps on capturing and learning complex relationships within graphs by making it effective in identifying anomalous patterns, introduced by [8]. GINs are predicated on the hypothesis that a neural network can learn to distinguish between different graph structures if it is as powerful as the Weisfeiler-Lehman (WL) graph isomorphism test, a classical algorithm for testing graph isomorphism. The key innovation of GINs lies in their update rule, which is designed to ensure that the representation of a graph is unique (up to isomorphism). Mathematically, the update rule for a node $v$ in GIN given by

$$h_v^{(k)} = MLP^{(k)}\left((1+\varepsilon^{(k)}) \cdot h_v^{(k-1)} + \sum_{u \in N(v)} h_u^{(k-1)}\right)$$

where;
- $h_v^{(k)}$ : The feature representation of node *v* at layer *k*.
- $MLP^{(k)}$ : A Multi-Layer Perceptron that is applied at the $k^{th}$ *layer.* It's a function that learns to map the input feature representation to a new feature space.
- $\varepsilon^{(k)}$ : A learnable parameter at layer *k* that can adjust the weighting of the node's previous features during the update.
- $h_v^{(k-1)}$ : The feature representation of node *v* at the previous layer $k-1$.
- $\sum(u \in N(v))$ : The aggregated feature representations of the neighbors of node *v* at the previous layer $k-1$.

Here, the presence of $h_u^{(k-1)}$ indicates that this is a recursive update, where the features from the previous layer are used to calculate the features for the current layer. However, the loss function utilized in GIN for tasks such as graph classification or node classification mainly similar to the above mentioned GNN types which involves the cross-entropy loss. As stated by [8], GIN is a straightforward example of one of the numerous maximally strong GNNs.

*d) GraphSAGE (Graph Sample and Aggregated) Networks:* [2] further refined this concept by introducing GraphSAGE, a method that extends GNNs to large-scale graphs through sampled neighborhood aggregation, demonstrating the scalability of message passing mechanisms. Unlike traditional approaches that require the entire graph to be loaded into memory, GraphSAGE learns a function to generate embeddings by sampling and aggregating features from a node's local neighborhood. GraphSAGE randomly samples a fixed number of neighbors for each node, reducing the computational complexity associated with large neighborhoods. It then aggregates the sampled neighbors' features using various functions (mean, LSTM, pooling) to generate a node's new feature representation. This method enables inductive learning, allowing GraphSAGE to generate embeddings for unseen nodes after training, making it highly effective for dynamic graphs where new nodes continually appear.

The architecture of the GraphSAGE model incorporates multiple convolution layers and leverages mean aggregation from neighbors to effectively encode node information within the graph structure, as demonstrated in Figure 10 of the source material [53]. The aggregation of features for a given node v is computed as the mean of the features of node *v*'s neighbors, mathematically expressed as:

$$h^i_{(N(v))} = \frac{1}{\|N(v)\|} D_p [h^{(i-1)}_u], \forall u \in N(v),$$

The aforementioned formula, $h^i_{(N(v))}$ denotes the aggregated feature representation of the neighbors of node *v* at the k-th layer. It should be noted that the forward pass through layer k may be incorporated into the computation of future nodes given by;

$$h^i_v = \sigma(concat[W^k_{self} D_p[h^{(i-1)}_v], W^i_{neigh} h^i_{(N(v))}] + b^k)$$

Here, $h^i_v$ signifies the updated feature representation after the concatenation of the node's own features with its neighbors', followed by the application of a non-linear transformation σ, with $W^k$ representing the weight matrix at layer *k* and $b^k$ the corresponding bias term. The concatenation operation introduces a structural mechanism to combine the features from the node itself and its neighbors, hence enriching the feature space. Dropout $D_p$ is applied as a regularizing measure during the aggregation process, introducing randomness with a probability p to mitigate the risk of over-fitting [53].

*E. Message Passing Framework*

The message passing framework, as introduced by [1], is a generalized approach that forms the backbone of various GNN architectures. It consists of two primary steps: aggregation and update. This framework enables the efficient processing of graph-structured data by allowing nodes to 'communicate' with each other through the exchange of messages, effectively capturing the structural dependencies within the graph.

*a) Message Passing with Time:* Extending the message passing framework to dynamic graphs involves incorporating temporal information into the aggregation and update processes. This extension is crucial for applications in fraud detection, where the relationships and behaviors of entities evolve over time. Models like Temporal Graph Networks (TGNs) introduced by [11], integrate time into the message passing framework, allowing for the capture of temporal patterns in node interactions and behaviors.

*b) Temporal Aggregation:* Within the context of dynamic graphs, messages are aggregated not just based on spatial proximity (neighborhoods) but also considering the temporal dynamics of interactions. This involves weighting messages by their recency or incorporating time stamps directly into the message function. Such an approach ensures that the aggregated information reflects the most relevant and recent interactions, providing a more accurate representation of the current state of the graph.

*c) Temporal Update:* The update function in a temporally aware message passing framework may also account for the time elapsed since the last update. This allows node representations to decay or evolve based on the freshness of the information. By incorporating time into the update process, the framework can model the dynamic nature of real-world graphs more effectively, where older interactions might be less indicative of the current state compared to more recent ones.

The integration of temporal information into the message passing framework marks a significant advancement in the field of graph neural networks, opening up new possibilities for analyzing and predicting behaviors in dynamic systems. By capturing both the structural and temporal dependencies within data, these models offer a powerful tool for a wide range of applications, from social network analysis to real-time fraud detection.

*F. Dynamic and Static Concepts in Fraud Detection Framework*

Recent research sheds light on the dynamic and static concepts within fraud detection frameworks, offering insights into their impacts, challenges, and advancements. This section synthesizes findings from several key studies to provide a comprehensive overview of current trends and future directions in fraud detection.

*a) Static Fraud Detection Frameworks:* Static fraud detection frameworks analyze historical data to identify patterns and anomalies to spot usual patterns or signs that might suggest fraud. These systems rely on predefined rules or models that do not change over time, making them well-suited for detecting known types of fraud. Nevertheless, these frameworks' static structure makes it more difficult for them to adjust to new and constantly changing fraudulent activities.

The primary limitation of static fraud detection frameworks is their inability to adapt to the rapidly changing tactics of fraudsters. As [4] highlight in their study on adaptive fraud detection using dynamic risk features, static models suffer from **"Concept Drift",** where changes in fraudulent patterns over time can significantly diminish the effectiveness of these models.

*b) Dynamic Fraud Detection Frameworks:* On the other hand, dynamic fraud detection frameworks are made to change in response to new data and emerging fraud patterns. These systems use deep learning, machine learning, and graph neural networks to update their models frequently based on the most recent data, which maintains the effectiveness over time.

Recent research has focused on enhancing the adaptability and effectiveness of dynamic fraud detection frameworks. [38] introduced a semi-supervised anomaly detection (SAD) framework for dynamic graphs, which combines a time-equipped memory bank and pseudo-label contrastive learning to efficiently uncover anomalies in evolving graph streams. This approach demonstrates the potential of dynamic frameworks to leverage large volumes of unlabeled data in detecting fraudulent activities.

Also, the need for real-time fraud detection in dynamic environments has led to the development of frameworks capable of identifying fraudulent communities within milliseconds. As a solution, [29] presented **Spade**, a real-time fraud detection framework for evolving graphs, which incrementally maintains dense sub-graphs to detect fraudulent communities in million-scale graphs. This framework highlights the importance of speed and scalability in combating fraud in highly dynamic settings.

## IV. EXISTING WORK

This section of the paper seeks to provide a summary of the most recent existing work related to the use of GNNs in the fraud detection framework, their model architecture and evaluate the results. However, there are no specific papers directly addressing the GNN approaches for fraud detection in the ride-hailing platforms which signifies the potential of this domain. Hence, the objective of this section is to assist the research community in enhancing current methods for detecting fraud or developing new strategies within the ride-hailing sector.

**Spatio-Temporal Attention Graph Neural Network (STAGN)** proposed by [31], a cutting-edge model that uses a spatial-temporal attention-based graph neural network to significantly improve the detection of credit card fraud, which can be used to analyze graph-structured data. This method cleverly combines the analysis of both time-related and geographical features from transaction data using a graph neural network. It then applies spatial-temporal attention mechanisms to precisely target and highlight patterns of fraud, revealing key areas and times where fraudulent activities are most likely to occur. The model processes raw transaction data through a series of steps, starting with a location-based GNN layer, then applying spatial-temporal attention, and finally using multiple 3D ConvNets to extract sophisticated representations of the data. These representations are then transformed into vectors and analyzed by a detection network to identify potential fraud. The entire process, from attention weight adjustment to 3D convolution and fraud detection, is optimized in a seamless, end-to-end manner, enhancing the model's ability to accurately predict fraud. By focusing on both the where and when of transactions, STAGN offers a nuanced approach to identifying fraud patterns, making it especially suitable for use in ride-hailing platforms where the geographical location and timing of services are crucial. In essence, STGCNs use convolutional neural networks (CNNs) to extract features from the spatial-temporal data and graph convolutional layers to capture the relationships between the data points. This allows the model to learn the spatial-temporal patterns in the data and make predictions or classify the data.

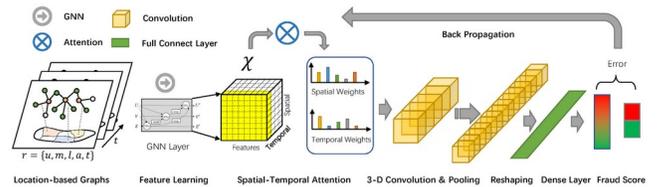

Fig. 6. STAGN Model Architecture

**Local-Global Mixing Graph Neural Network ( LGM-GNN)** model introduces a novel approach to fraud detection by harmonizing local and global information through a memory-based Graph Neural Network. This approach starts by understanding the details of different interactions through relation-aware embedding, followed by an aggregation process that combines both local and global insights to identify fraudulent activities effectively. At its core, the architecture leverages memory networks to store, refine, and update relevant information, enabling the model to counteract fraudster camouflage and noise. A hierarchical information aggregator further enhances the model's capability by combining refined embeddings for final decision-making. This innovative approach addresses key challenges such as class imbalance, fraudulent camouflage, and context inconsistency, which are challenges that traditional machine learning models struggle to handle effectively.

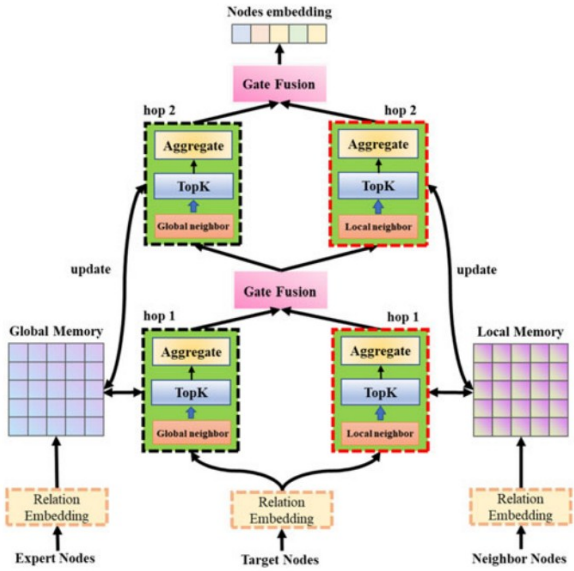

Fig. 7. LGM-GNN Structure

This model offers a promising solution for detecting fraud in online ride-hailing platforms by masterfully analyzing the complex web of user interactions. By representing passengers, drivers, and trip details as nodes within a graph, the model utilizes relation-aware embeddings to highlight unusual patterns, such as abnormal ride frequencies or payment methods, which could indicate fraudulent behavior. Its capacity to aggregate both local and global information allows for the detection of sophisticated fraud schemes, including those involving multiple accounts or coordinated actions across the network. The incorporation of memory networks enables the system to recognize and adapt to evolving fraud tactics over time, enhancing its predictive accuracy. Additionally, the models' hierarchical information aggregator ensures comprehensive analysis, from individual behaviors to overarching patterns, thereby identifying fraudulent activities with greater precision.

**Multi-view Similarity-based Graph Convolutional Network (MSGCN)** model introduced by [25], which is proficient at navigating through the complex user interactions typical of ride-hailing platforms. This model shines in its ability to sift through the dense and varied data, pinpointing unusual or potentially fraudulent behavior by tapping into the rich, layered connections found within heterogeneous information networks (HINs). By crafting a detailed graph that maps out the myriad entities and their interactions, MSGCN employs meta-path techniques to break down this complex graph into simpler, single-view graphs. Each of these focuses on distinct relationship types, allowing the model to grasp the subtle nuances in node relationships that are key to spotting fraud. The model employs a similarity-based approach within the GCN to learn deep insights about node representations, focusing on both their structural and semantic similarities. This is crucial for identifying patterns that might indicate fraudulent activities. To ensure that the most pertinent information is highlighted, an attention mechanism is used to weigh these features, fine-tuning the model's focus for sharper fraud detection accuracy. The effectiveness of this method has been proven in experiments, where it outshines traditional fraud detection techniques, showcasing its potential as a formidable tool against fraud in online environments.

This approach outlines a strategy for identifying fraud within a Heterogeneous Information Network (HIN) using a GCN. Initially, the HIN, characterized by its complex web of relationships among different types of entities, is broken down into simpler, single-view graphs. Each of these graphs focuses on particular patterns of interaction. The GCN then analyzes these graphs to identify key node features, paying special attention to both the structural and semantic similarities that emerge. To synthesize the insights gained from each graph, an attention mechanism is employed, ensuring that the most significant features for fraud detection are highlighted. The node representations produced through this process are rich and multifaceted, capturing the essence of the network's interactions. Ultimately, the method assigns a binary classification to each node, signaling the presence or absence of fraudulent activity. This technique proves to be highly effective in sifting through complex information to pinpoint fraud, thereby safeguarding the integrity of digital platforms.

The adaptability of the MSGCN is particularly key for ride-hailing platforms, given the ever-evolving nature of fraudulent schemes. The model should be able to quickly spot potential fraud by analyzing different types of data. It helps keep the platform safe and trustworthy by finding and focusing on unusual patterns in trip rides, drivers, or passengers. This way, ride-hailing services can take early action to prevent fraud and make sure their services remain secure and reliable.

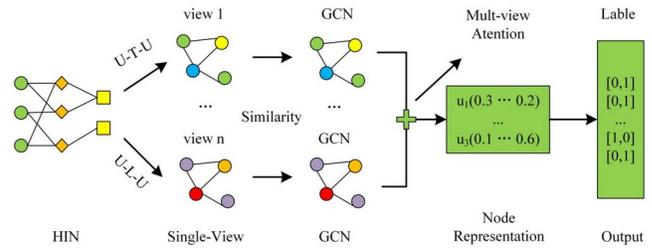

Fig. 8. MSGCN Architecture

## V. ANALYSIS OF EXISTING WORK

This section of the paper analyses and comapre the main three solutions under features and characteristics.

TABLE I. COMPARATIVE ANALYSIS OF GRAPH NEURAL NETWORK MODELS FOR FRAUD DETECTION

| *Feature* | **STAGN** | **LGM-GNN** | **MSGCN** |
|---|---|---|---|
| *Dataset* | Credit card transactions | Amazon, YelpChi | MicroblogPCU from Sina Weibo |
| *Key Techniques* | Spatial-temporal attention, 3D convolution | Local and global information integration, memory-based GNN | Multi-view graph convolution, Meta-path-based user associations |
| *Batch Size* | Not specified | 256 (Amazon), 1024 (YelpChi) | Not specified |
| *Optimizer* | Not specified | Adam | Not specified |
| *Learning Rate* | Not specified | 0.01 (Amazon), 0.001 (YelpChi) | Not specified |

## VI. SUMMARIZED TAXONOMY

In the Summarized Taxonomy section, the author provides a structured overview of GNN architectures and corresponding methodologies as they apply to anomaly detection, with a special emphasis on fraud detection—a critical concern within ride-hailing platforms. This classification not only differentiates the various anomaly types and associated graph structures but also captures significant methodological progress addressing particular challenges in this field.

TABLE 2. COMPREHENSIVE TAXONOMY OF GNN ARCHITECTURES FOR ANOMALY DETECTION

| Graph Type | Anomaly | Architecture | Method | Summary (Key issue →Solution) |
|---|---|---|---|---|
| Static Graph | Node Level | GCN-based GAE | DOMINANT [10] | Complex interactions, sparsity, non-linearity → GCN-based encoder |
| | | | ConsE [46] | Sparsely labeled graphs using GNNs prone to under-fitting→ Semi-supervised GCN-based model |
| | | GAT-based GAE | AnomalyDAE [16] | Complex interactions → GAT-based encoder |
| | | | AEGIS [17] | Handling unseen nodes → generative adversarial learning with GAE |
| | | GCN | Semi-GCN [19] | Label information → semi-supervised learning by GCN |
| | | | MSGCN [25] | Fraud detection → detecting anomalies in heterogeneous social networks by utilizing a multi-view similarity-based graph convolutional network |
| | | Other GNN-based models | GraphBERT [36] | Relationships between user and speeches in social networks→ Integrating tweet and user representation as node features |
| | | | LGM-GNN [50] | Fraud detection → graph-based fraud detection integrating local and global information using a memory-based approach. |
| | Edge Level | GCN-based GAE | GFDN [47] | Group based fraud detection → GDFN |
| | | GCN | SubGNN [26] | Fraud detection→ GIN and extracting and relabeling subgraphs |
| | Sub-Graph-Level | GAT-based GAE | HO-GAT [27] | Abnormal sub-graphs → hybrid-order attention with motif instances |
| | | GCN | GDFN [37] | Non utilization of e-commerce metadata → fusion technique to synthesize low- and high-order interactions with sub-graphs |
| | | Other GNN-based models | STAGN [31] | Domain Knowledge needed for feature generation in transactions→ Utilizing temporal and location-based transaction graph features |
| Dynamic Graph | Graph Level | GAT-based GAE | DEDGAT [48] | Directional bias in existing GAT models→Explicitly calculates in-degree and out-degree representations |
| | | | GLAM [33] | Addressing Distribution anomalies → MMD Pooling |
| | Node Level | GCN & GRU | DEGCN [34] | To capture node- and global-level patterns → DGCN & GGRU |
| | Edge Level | GCN & GRU | Hierarchical-GCN [18] | Dynamic data evaluation → temporal & hierarchical GCN |
| | Sub-Graph Level | GCN-based GAE | BitcoNN [49] | Class imbalance, and large-scale data→Adaptive neighbor sampling |
| | | Other GNN-based models | DyHGN [35] | Dealing with graphs that are both heterogeneous and dynamic → Capture both temporal and heterogeneous information |
| | Graph Level | Other GNN-based models | GADY [51] | Dynamic structure constructing and negative sampling→ Message-passing with positional features and GANs to generate negative interactions. |

## VII. RESULTS AND DISCUSSION

The expansion of digital transactions and ride-hailing platforms has led to a spike in fraudulent activity, especially after the COVID-19 pandemic.Traditional fraud detection methods, relying on statistical models and machine learning algorithms, fall short in capturing the complex and dynamic interactions within ride-hailing ecosystems. Hence, the emergence of GNNs marks a significant change, via relational data to more accurately detect fraudulent activities through analyzing the data from multiple perspectives to uncover new, complex relationships between the entities.

This paper analyzes popular fraudulent activities within the ride-hailing ecosystem and evaluates the effectiveness of various GNN models in detecting these frauds. It explores the intersection of fraud detection with GNNs, addressing both dynamic and static concepts in fraud detection frameworks.

Globally, several critical fraudulent variations are prevalent in ride-hailing platforms, including fake GPS systems, route manipulation, long-hauling, GPS spoofing, ride collusion, and hire conversions/trip manipulations. However, it is noted that there are no relatable studies addressing hire conversions and trip manipulations, even though these are critical fraudulent incidents that affect both organizations and passengers. Thus, the author identifies a significant research gap in this specific area of study.

Existing studies have primarily focused on models like the STAGN, LGM-GNN, and MSGCN, each addressing different aspects of fraud detection with the potential for integration into the ride-hailing industry. Among these, the LGM-GNN stands out due to its innovative approach in integrating local and global information through a memory-based mechanism. This approach effectively addresses key challenges such as class imbalance and fraudulent camouflage, making LGM-GNN particularly suited for fraud detection frameworks in the ride-hailing industry. Its ability to identify complex patterns and anomalies within large-scale, LGM-GNN models can better spot fraud scenarios within the complex data, making it highly effective for tracking advanced fraud techniques in the industry.

To assist the research community in identifying potential research gaps and current techniques, the summarized taxonomy section offers a structured overview of GNN architectures and methodologies applied to anomaly detection, especially in fraud detection. It highlights significant methodological progress and differentiates anomaly types and associated graph structures, offering insights into the evolution of fraud detection strategies in ride-hailing platforms.

While the research into fraud detection using GNNs shows promise, there are significant gaps needing improvement. Particularly, limited exploration of how these models perform in the real world. Also, some technical issues such as the risk of overfitting in models that do not sufficiently address class imbalance or overlook the temporal dynamics of fraud patterns, which can severely undermine their effectiveness in detecting and preventing fraud in such a dynamic and fast-paced industry.

## VIII. CONCLUSION

The paper offers a thorough examination of fraud detection within the ride-hailing industry through the lens of GNNs. It meticulously analyzes various fraudulent activities and evaluates the efficacy of different GNN models in addressing these challenges. Highlighting innovative GNN models and for its superior ability to integrate local and global information, the study underscores the various types of model effectiveness in combating class imbalance and fraudulent camouflage. Furthermore, it provides a structured overview of GNN architectures and methodologies, shedding light on significant methodological progress. Despite advancements, the paper identifies gaps in real-world applicability and technical shortcomings, such as potential overfitting and the neglect of temporal dynamics in fraud patterns and specially there are potential use cases, such as the detection of fraudulent trips and the identification of relevant drivers or passengers. These applications could greatly benefit from integrating big data concepts with GNNs, by enhancing the ability to effectively process and analyze complex data patterns. This comprehensive analysis not only contributes to the academic discourse on fraud detection within the online ride-hailing platforms but also opens avenues for future research to refine and adapt GNN-based methods for enhanced fraud detection capabilities.

ACKNOWLEDGMENT

With great appreciation, the authors would like to thank Hamdaan Mohideen for his vital contributions to this study. The direction and success of the study have been greatly influenced by his knowledge, viewpoints, and suggestions. His assistance improved the caliber of the research findings as well as the research process. We have the utmost gratitude for his commitment and the vital roles he played in making this study possible.